\documentclass[10pt,twocolumn,letterpaper]{article}

\usepackage[pagenumbers]{cvpr} 

\usepackage[utf8]{inputenc} 
\usepackage[T1]{fontenc}    
\usepackage{url}            
\usepackage{booktabs}       
\usepackage{amsfonts}       
\usepackage{nicefrac}       
\usepackage{xcolor}         
\usepackage{amsmath}        
\usepackage{algorithm}      
\usepackage{algpseudocode}
\usepackage{graphicx}
\usepackage{mdframed}
\usepackage{xcolor}
\usepackage{multirow}
\usepackage{enumitem}

\usepackage{textcomp}
\usepackage[most]{tcolorbox}
\lstdefinestyle{prompt}{basicstyle=\ttfamily\scriptsize, breaklines=true, breakatwhitespace=true,
  columns=fullflexible, keepspaces=true, showstringspaces=false,
  literate={'}{\textquotesingle}1, breakindent=0pt, aboveskip=0pt, belowskip=0pt,
  moredelim=[is][\color{cvprblue}]{|}{|}}
\newtcblisting{promptbox}[1]{
  enhanced, breakable, listing only,
  colback=gray!4, colframe=gray!55, boxrule=0.5pt, arc=2pt,
  left=4pt, right=4pt, top=3pt, bottom=3pt,
  title={#1}, fonttitle=\bfseries\small, coltitle=black, colbacktitle=gray!15,
  listing options={style=prompt}
}

\usepackage{microtype}

\renewcommand{\paragraph}[1]{\smallskip\noindent\textbf{#1}}

\AtBeginEnvironment{equation}{\small}
\AtBeginEnvironment{align}{\small}
\AtBeginEnvironment{gather}{\small}
\AtEndEnvironment{equation}{\vspace{-6pt}}
\AtEndEnvironment{align}{\vspace{-6pt}}
\AtEndEnvironment{gather}{\vspace{-6pt}}

\DeclareMathOperator*{\argmax}{arg\,max}

\usepackage{soul}

\definecolor{cvprblue}{rgb}{0.21,0.49,0.74}
\usepackage[pagebackref,breaklinks,colorlinks,allcolors=cvprblue]{hyperref}

\def\confName{CVPR}
\def\confYear{2027}

\title{Stress-Testing Visual Recognition and Reasoning in the Goldilocks Zone}

\author{
Naren Sivakumar \qquad Frank Ferraro \qquad Tejas Gokhale\\
University of Maryland, Baltimore County\\
{\tt\small \{narens1, ferraro, gokhale\}@umbc.edu}
\vspace{-10pt}
}
\begin{document}

\twocolumn[{%
  \renewcommand\twocolumn[1][]{#1}%
  \maketitle

  \begin{center}
    \centering
    \includegraphics[width=\linewidth]{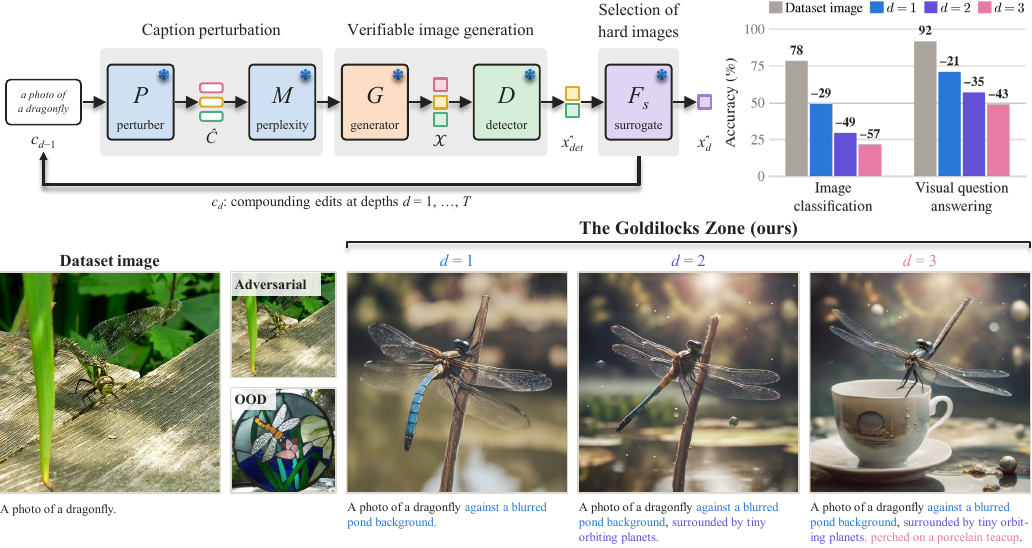}

    \captionof{figure}{
      We propose a method for automated generation of stress-test datasets from an existing benchmark dataset. 
      Our approach leverages caption perturbations within a perplexity range and verifiable image generation that preserves object presence.
      Compared to current paradigms of adversarial (small pixel-level perturbations) and out-of-distribution (OOD) robustness (large distribution shifts), our stress tests are in the ``Goldilocks Zone'' between these two extremes.   
      This approach can operate at several depths $d$ by bootstrapping perturbations and generating progressively more challenging test cases.
      Our results demonstrate that on multiple image classification and visual question benchmarks, Goldilocks zone stress-testing reveals failures of vision and vision-language models, and opens a new way to stress-test them.
    }
    \label{fig:teaser}
  \end{center}
}]
  
\begin{abstract}
\vspace{-4pt}

Evaluation of model robustness operates at two extremes: robustness to small perturbations such as imperceptible adversarial attacks or corruptions and out-of-distribution generalization under large dataset-level shifts. 
We propose a new paradigm -- the ``Goldilocks Zone'' -- that operates between these extremes. 
Our method starts with test images from existing benchmark datasets, distills them into their core semantics using captioning, and searches for 
caption perturbations within a perplexity range as a proxy for linguistic plausibility. Using text-to-image generators with these optimally perturbed captions as input, we obtain new datasets containing challenging images to stress-test vision models. 
By bootstrapping previous edits and compounding additional edits, our approach allows generation of stress-test datasets at multiple depths, thus posing increasing difficulty for downstream models.
Comprehensive experiments demonstrate the challenge posed by our automatically constructed stress-tests on visual recognition and reasoning tasks. 
State-of-the-art models struggle on the Goldilocks stress-tests, with their accuracy of vision models decreasing by $\sim45\%$ on image classification and that of vision-language models by $\sim23\%$ on visual question answering.
\end{abstract}

\section{Introduction}

Vision models achieve excellent performance across several tasks ranging from image classification~\citep{He_2016_CVPR,liu2022convnet} to visual question answering \cite{team2026gemma, qwen36_35b_a3b}.
However, they fail to generalize outside their training distribution, thus calling for comprehensive evaluation of their robustness.
Existing approaches include adversarial attacks to test model performance against small pixel-level perturbations \cite{carlini2017towards, Dong_2018_CVPR}, natural corruptions \cite{hendrycks2018benchmarking}, out-of-distribution (OOD) domains \cite{recht2019imagenet,hendrycks2021natural}, or spurious correlations \cite{sagawa2020distributionally}.
Recent work has explored synthetic generation of \textit{counterfactual} images~\citep{prabhu2023lance, zhang2024imagenet} for isolating attributes that cause model failures and vulnerabilities.
However, they are limited to two extremes -- small pixel-level perturbations or large distribution-level shifts. A \emph{stress test} pushes a system beyond its normal operating conditions to probe and expose weaknesses, while still meeting its intended specification~\citep{naik-etal-2018-stress}. 
For a vision model, this means images that differ visibly and semantically from its test set, yet stay within its domain and keep their label, which neither extreme provides.

In this work, we aim to leverage the middle ground between the two extreme robustness paradigms, with our method \textit{The Goldilocks Zone}\footnote{Not too hot, not too cold, just right!~\citep{southey2013goldilocks}}, a fully automated, language-guided stress-testing framework. 
Given a labeled dataset of images, our method distills the images into their core semantics using captioning, and searches for perturbations that are linguistically plausible.
Our framework utilizes two modules (caption perturbation and image generation) that perform edits while ensuring the retention of the ground truth label to create stress-testing images. 
We ensure that the proposed edits are plausible through two checks: (1) a perplexity filter, which acts as a proxy for linguistic plausibility, and (2) a verifier object detector, that ensures that the true label is still in the newly created stress-testing image. 
We bootstrap our findings across multiple iterations, or depths, of this process -- edits that cause the maximum loss in a downstream model are preserved and improved upon in later depths, leading to progressively more challenging datasets. 

Our experiments reveal several interesting findings. 
First, we show that stress-testing in the \textit{Goldilocks Zone} leads to significant accuracy drops of $\sim45\%$ on multiple image classification benchmarks, exceeding existing stress-tests and OOD datasets. 
Second, we find that modern VLMs also fail in a similar fashion on the visual question answering task, with accuracy drops of up to $\sim41\%$ on Goldilocks-generated images while outperforming most VQA stress-test baselines.
Third, we show that there are shared vulnerabilities among different downstream models that could be exploited to generate difficult stress-testing images.  

Our comprehensive experimental results and analyses demonstrate the efficacy of stress-testing in the Goldilocks zone and point to a new evaluation paradigm in which test samples are obtained via automated generation, without human involvement, and by probing model failures for dynamic and progressive model evaluation.

\begin{figure*}[t]
    \centering
    \includegraphics[width=0.95\linewidth]{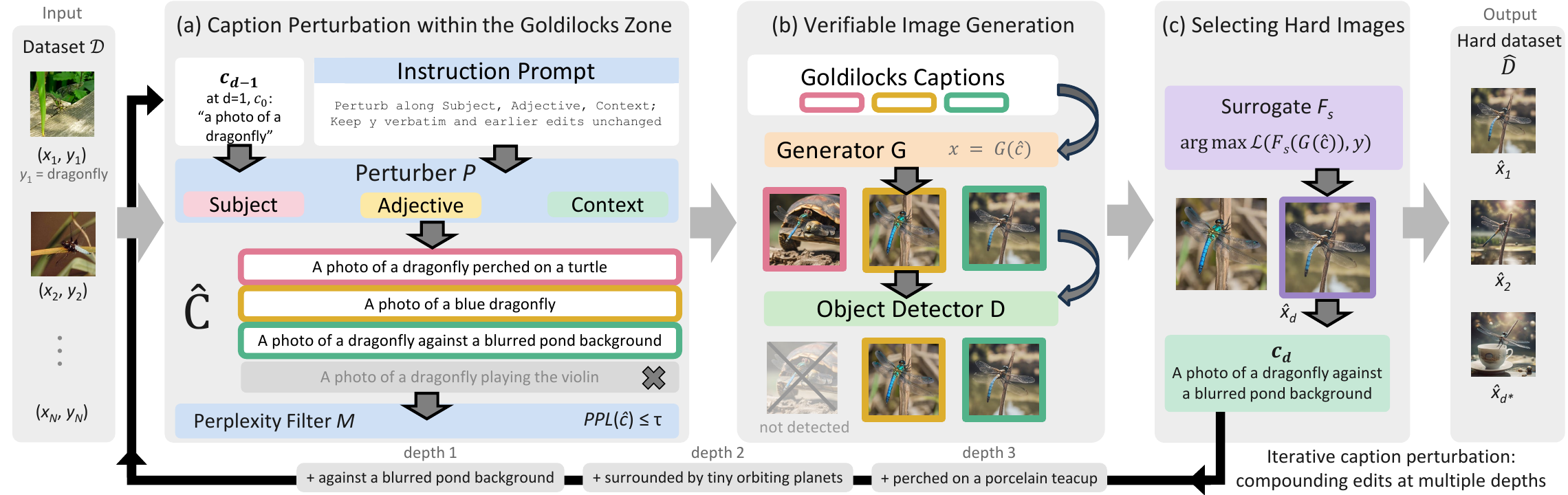}
    \caption{The Goldilocks Zone: Given a template caption of an image from a dataset with the true label of the image, we edit the caption along specific linguistic axes, and only retain captions that pass through a perplexity filter. These captions are  generated into images through a text-to-image generator, and verified by an object detector to ensure the true label is preserved. The surviving images are then passed to the surrogate vision model, and the caption whose image caused the most loss in the surrogate is then bootstrapped for the next iteration. 
    }
    \label{fig:methodfig}
\end{figure*}
\section{Related Work}

\paragraph{Adversarial robustness in vision} 
has been studied in the context of pixel-level attacks that find a small bounded perturbation optimized to increase model failures~\citep{DBLP:journals/corr/GoodfellowSS14, carlini2017towards}.
Adversarial training~\citep{salman2020adversarially, madry2018towards} serves as the primary approach to mitigate adversarial attacks.
Several variants of adversarial attacks have been developed such as FGSM, PGD, and C\&W~\citep{DBLP:journals/corr/GoodfellowSS14, carlini2017towards, madry2018towards} along with standardized benchmarks such as RobustBench~\citep{DBLP:conf/nips/CroceASDFCM021}. 
In contrast to imperceptible adversarial attacks, our work generates perceptible semantic perturbations of images as a stress-test for models.

\paragraph{Distributional robustness}
evaluation focuses on large shifts from the training domain~\citep{hendrycks2021natural, wang2019learning, barbu2019objectnet, li2017deeper, pmlr-v139-koh21a, recht2019imagenet}. 
Prior work on improving distributional robustness has leveraged data augmentation~\citep{hendrycks*2020augmix, hendrycks2021many}, domain invariance~\citep{ganin2016domain, arjovsky2019invariant}, and group robustness strategies~\citep{sagawa2020distributionally, liu2021just}.
Our approach is to remain closer to the training domain, but to explore semantics in the long tail of the domain, 
by dynamically generating stress-testing images per input. 
Our contribution is not a new dataset but rather an automated algorithmic framework for generating new challenging datasets from existing ones, thus making our approach extensible and plug-and-play.

\paragraph{Diffusion models for data augmentation}
have been used in two forms; with original images either being edited~\citep{trabucco2024effective, dunlap2023diversify, islam2024diffusemix, meng2022sdedit} or regenerated from captions, class labels or the original images~\citep{azizi2023synthetic, tian2023stablerep, yuan2024realfake, shipard2023diversity, sariyildiz2023fake}. 
Regeneration-based methods use the synthetic images to train models, such as, 
using an ImageNet fine-tuned T2I model to generate additional training data~\citep{azizi2023synthetic},
prompting a T2I model with class names to build a synthetic ImageNet for model training from scratch~\citep{sariyildiz2023fake},
or treating several images generated from the same caption as positive augmentations for contrastive representation learning~\citep{tian2023stablerep}. 
Unlike these methods, we regenerate images to evaluate models: we search over compounding caption edits for challenging images 
and automatically generate a stress-test dataset given a seed dataset. 
    
\paragraph{Language Guided Counterfactuals.} 
Prabhu \etal~\citep{prabhu2023lance} and Luo \etal~\citep{luo2023zero} generate counterfactual images by swapping objects, attributes, and contexts to reveal model vulnerabilities, demonstrating that models remain brittle under semantically meaningful modifications. 
However, rule-based caption edits and simple word swaps often produce non-fluent or nonsensical text~\citep{hsieh2023sugarcrepe, morris2020reevaluating}, thus obfuscating failures due to semantics with failures due to undesirable artifacts, such as 
models exploiting implausible text to appear more capable~\citep{hsieh2023sugarcrepe}, or failing on invalid inputs, inflating failure rates~\citep{morris2020reevaluating}. 
This creates a need for a plausibility constraint, 
which we address 
through iterative optimization under a perplexity-based filter, encouraging perturbations to remain linguistically natural and visually coherent.
\vspace{-4pt}
\section{Methodology}
\vspace{-4pt}
\label{sec:method}
We generate challenging test datasets in the Goldilocks Zone, through semantic perturbations that are large enough to explore diverse regions in the distribution, yet constrained enough to preserve linguistic plausibility and target-label validity. 
Given an existing test dataset $\mathcal{D}$, our method produces its Goldilocks version $\hat{\mathcal{D}}$.
Our method achieves this through an LLM-based perturber $P$ that edits the natural language description of an image within a perplexity range, a text-to-image diffusion model that generates images corresponding to the perturbed captions, followed by selection of maximally challenging images using a weak surrogate model $F$.
\cref{fig:methodfig} shows an overview and we describe it in detail below. Pseudo-code is in supplement material.

\subsection{The Goldilocks Zone Pipeline} 
Given an image $x$ with ground-truth label $y$, our goal is to \textit{generate} synthetic images $\hat{x}$ to stress-test a vision model.

\paragraph{Caption Perturbation Within The Goldilocks Zone.} 
The initial caption $c_0$ is a template that contains true label $y$; for \eg, ``a photo of a $dragonfly$'' in \cref{fig:methodfig} (a).  
We use a large language model (LLM) to perturb this caption along three axes, namely: \emph{Subject}, that introduces another entity that interacts with or occupies the scene alongside the object; \emph{Adjective} that modifies descriptive attributes; and \emph{Context} that changes background or scene, as shown in \cref{fig:methodfig}~(a).
These axes are chosen to reflect the failure modes identified in robustness literature: failures due to co-occurring and out-of-context objects~\cite{singla2022salient}, over-reliance on texture and other surface attributes of the object~\cite{geirhos2018imagenettrained}, and using the background as a shortcut~\cite{beery2018recognition, sagawa2020distributionally, 11375645}.
We ensure (through the prompt to the LLM as well as word-level sanity checks) that the perturbations do not affect the ground-truth label $y$ in the caption.
We also prohibit the LLM from adding perturbations that would change the style of the described images (such as photo to sketch or cartoon) which would make the generated image out-of-distribution -- a case already covered by domain generalization benchmarks \cite{venkateswara2017deep,li2017deeper,peng2019moment}.

Once these perturbed captions $\mathcal{\hat{C}}$ are obtained, we develop a perplexity-based filter that is used to discard the implausible or linguistically incoherent captions. 
Perplexity has been used as a measure of fluency~\citep{bengio2003neural, radford2019language, brown2020language}, a filtering mechanism for atypical samples~\citep{axelrod2011domain, moore-lewis-2010-intelligent}, and a proxy for semantic plausibility~\citep{yuanprobelm, xu2025perplexity}, making it suited as our filter.
Captions above perplexity threshold $\tau$ are discarded, thus giving us the Goldilocks Zone captions:
\begin{equation}
    \mathcal{C}_\mathrm{Goldilocks} = \{ \hat{c} \in {\mathcal{\hat{C}}} \mid \mathrm{PPL}(\hat{c}) < \tau \}.
    \label{eq:goldilocks_captions}
\end{equation}

\paragraph{Verifiable Image Generation.}
We use a text-to-image diffusion model $G$ to produce images $\mathcal{X}$ from our Goldilocks captions $\mathcal{C}_\mathrm{Goldilocks}$ (one image per caption).
To ensure that each generated image contains the ground-truth $y$, we use an oracle object detector $D$.
Images in which $y$ is not detected are discarded; for \eg , as shown in \cref{fig:methodfig}~(b), the dragonfly is less visible against the turtle, and is therefore rejected by the object detector. 
We thus obtain a verified image set:
\begin{equation}
    \mathcal{X}_\mathrm{verified} = \{x \in \mathcal{X} \mid y \in D(x)\}.
    \label{eq:verified_images}
\end{equation}

\paragraph{Selection of Hard Images.} 
Next, we select the most challenging images from $\mathcal{X}_\mathrm{verified}$ as shown in \cref{fig:methodfig}~(c).
To do this, we use a surrogate model, $F_s$, trained for the task being evaluated, and pass images from $\mathcal{X}_\mathrm{verified}$ through the surrogate model and record the loss w.r.t.\ ground-truth label $y$.
The maximally challenging images $\hat{x}^\ast$ and the corresponding captions $\hat{c}^\ast$ are obtained as:
\begin{equation}
    \hat{x}^{\ast} = \argmax_{x \in \mathcal{X}_{\mathrm{verified}}} \mathcal{L}\big(F_s(x), y\big), 
    ~~\textit{where}~~\hat{x}^\ast = G(\hat{c}^\ast).
    \label{eq:mainobj}
\end{equation}

\paragraph{Bootstrapping.} 
The above process represents one forward pass of perturbing the caption and generating a challenging image.
We bootstrap this approach by compounding the caption perturbations for multiple depths $d \leq T$ where $T$ is the maximum number of depths.
The selected caption $\hat{c}^{\ast}$ at depth $d$ becomes $c_d$, its added element joins the set of locked edits $A$, and $P$ proposes the next edits from $c_d$ under the instruction to preserve every edit in $A$. For example, in \cref{fig:methodfig}~(c), the perturbation ``against a blurred pond background'' is the most challenging to $F_s$, and is preserved by $P$ while proposing the new edit, ``surrounded by tiny orbiting planets''. We repeat this search for up to $T$ depths, with edits at each depth compounding. In \cref{fig:methodfig}, the depth 3 caption is the culmination of each step before it: ``A photo of a dragonfly against a blurred pond background\textsubscript{1}, surrounded by tiny orbiting planets\textsubscript{2}, perched on a porcelain teacup\textsubscript{3}.''. This bootstrapping process allows us to generate more challenging stress-testing images the further we go, while language \cref{fig:methodfig}~(a) and image checks \cref{fig:methodfig}~(b) at each depth ensure that the proposed edits are within the realm of plausibility.

\begin{figure*}[t]
    \centering
    \includegraphics[width=\linewidth,trim={0 6 0 0}, clip]{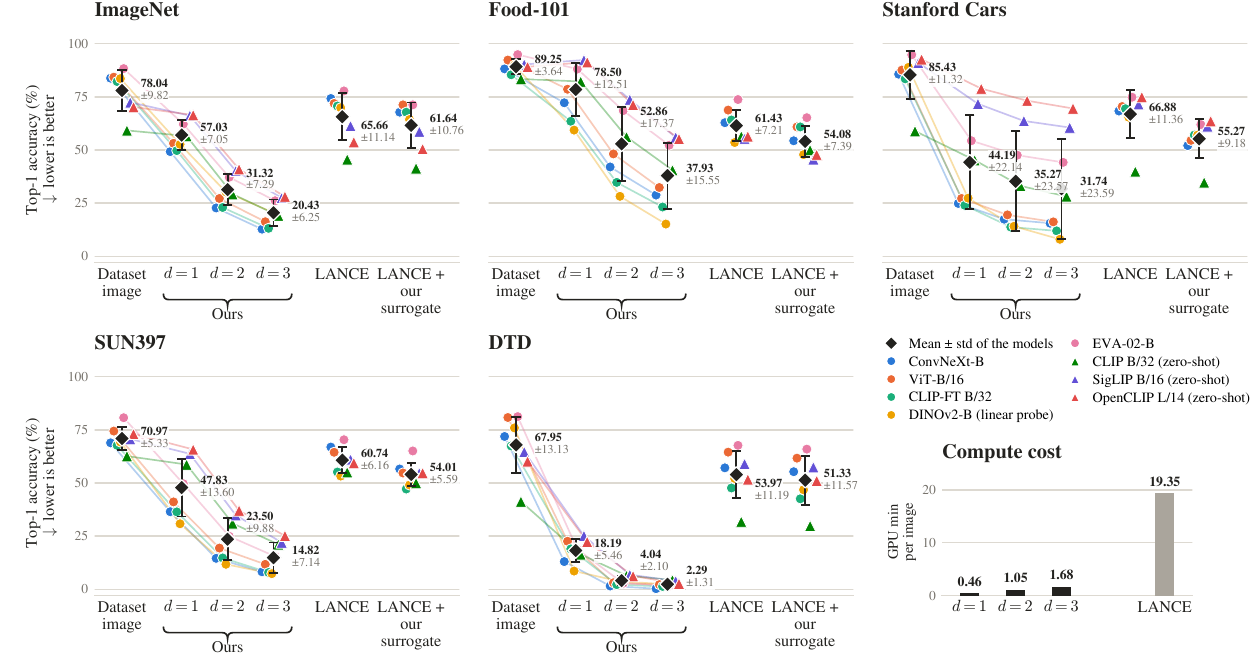}
    \caption{
    Comparing the performance of image classification models on existing benchmark datasets with LANCE \cite{prabhu2023lance} and our Goldilocks Zone test sets generated at depths $d=1, 2, 3$.
    We use ResNet-50 as a weak surrogate model and test the performance of 8 stronger SOTA target models (circles: supervised; triangles: zero shot). 
    Goldilocks datasets are challenging for all target models on each benchmark, with accuracy declining at higher depth of the perturbation.
    Our approach takes significantly less GPU minutes per image for image generation.
    }
    \label{fig:main_results}
\end{figure*}

\begin{table}
\caption{Number of images in the source test datasets and our generated stress-test dataset for image classification (surrogate: ResNet-50) and visual question answering (surrogate: Gemma-4 12B). Our automated construction of images enables a comprehensive evaluation beyond the in-domain data}
\centering
\small
\resizebox{\linewidth}{!}{
\begin{tabular}{@{}llrrrrr@{}}
\toprule
Task & Dataset & $|\mathcal{D}|$ & $|\hat{\mathcal{D}}|$ & $|\hat{\mathcal{D}}_1|$ & $|\hat{\mathcal{D}}_2|$ & $|\hat{\mathcal{D}}_3|$ \\
\midrule
\multirow{5}{*}{\shortstack[l]{Image\\ Classification}}  
    & ImageNet      & 9,950  & 178,170 & 54,008 & 62,298  & 61,864  \\
    & Food-101      & 1,010  & 20,546  & 6,606  & 7,194   & 6,746   \\
    & Stanford Cars & 1,960  & 52,318  & 17,338 & 17,332  & 17,648  \\
    & SUN397        & 3,870  & 59,645  & 15,229 & 22,083  & 22,333  \\
    & DTD           & 470    & 10,475  & 1,860  & 4,059   & 4,556   \\
\midrule
\multirow{2}{*}{\shortstack[l]{Visual Question \\Answering}} 
    & VQAv2         & 6,122  & 122,545 & 40,577 & 41,610  & 40,358  \\
    & GQA           & 17,675 & 274,844 & 70,777 & 100,048 & 104,019 \\
\bottomrule
\end{tabular}
}
\label{tab:sizes}
\end{table}
Upon obtaining stress-testing datasets at depths $d\in\{1,\dots T\}$, we use them for comprehensively stress-testing different downstream tasks and models.
\vspace{-4pt}
\section{Experiments}
\label{sec:experiments}
\vspace{-4pt}
\subsection{Experimental Settings}

\textbf{Implementation Details for Goldilocks.}
We use GPT-6 Luna \citep{openai_gpt6_luna} as the caption perturber, configured for structured JSON outputs to minimize parsing errors and unwanted strings in the perturbed captions. 
We obtain perplexity values from Llama 3.1-8B \citep{grattafiori2024llama} language model for the perplexity filter with threshold $\tau$ = 80. 
We use SDXL \citep{podellsdxl} as the text-to-image (T2I) model to generate images, and Owl-ViT~\citep{minderer2022simple} as our object detector.
All code is written in PyTorch and experiments use Nvidia L40S and H100 GPUs.

\begin{figure*}[t]
    \centering
    \includegraphics[width=\linewidth,trim={0 6 0 0}, clip]{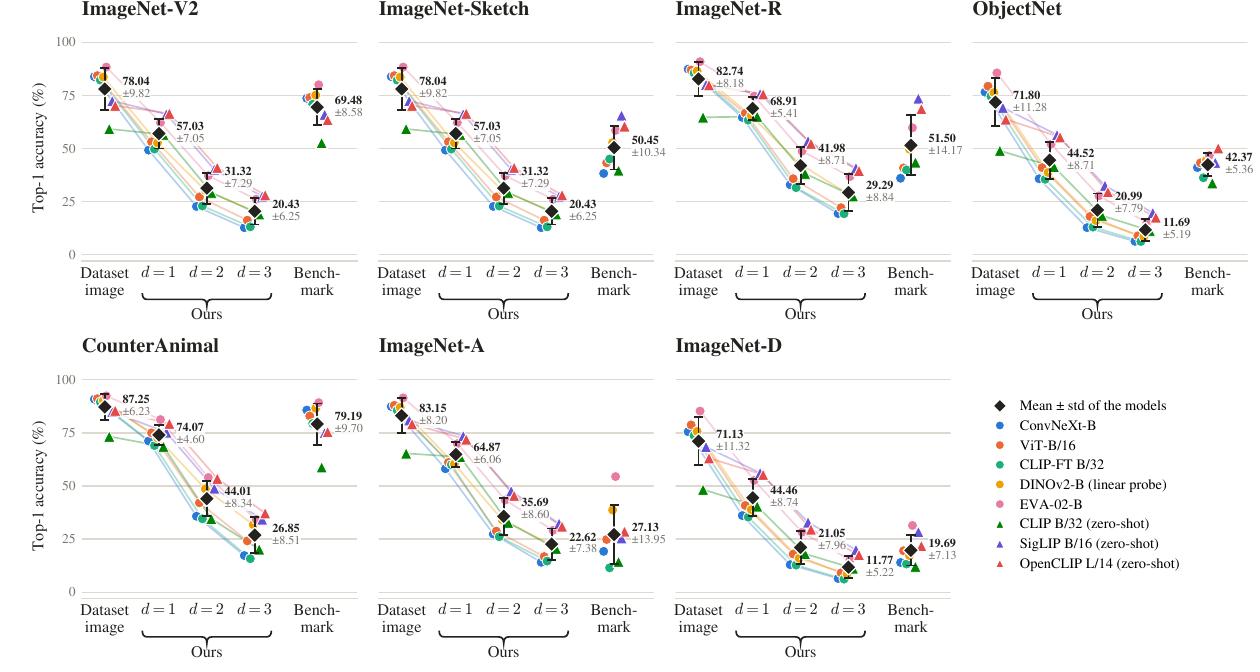}
    \caption{We compare our dataset generated for ImageNet against other datasets that use ImageNet as their seed, and use some human effort to curate the evaluation dataset. For parity, we evaluate against each dataset using only the classes they have released. 
    Our method outperforms the stress-testing datasets in comparison. The real photo column refers to the performance of the downstream model performance on the specific classes for each benchmark; the benchmark column gives how those models perform on the method's released dataset.}
    \label{fig:human_comp}
\end{figure*}

\paragraph{Datasets.} For \ul{image classification}, we use ImageNet \cite{deng2009imagenet}, Food-101 \cite{bossard14}, Stanford Cars \cite{6755945}, SUN397 \cite{xiao2010sun}, DTD \cite{cimpoi2014describing} encompassing diverse categories like objects, food, cars, scenes, and textures. For \ul{visual question answering} (VQA) we use the VQAv2 \cite{goyal2017making} and GQA \cite{hudson2019gqa} datasets that necessitate several types of reasoning.
For the template $c_0$, we use the CLIP zero-shot template, ``a photo of a $y$'' for each dataset. 
For VQA specifically, to avoid answers to questions being removed during the generation process, we only retain data points that satisfy the following criteria:
We keep a question if 
(1) its answer $y$ is a physical object, so that the detector $D$ can confirm $y$ in every edited image; 
(2) at least 8 of the 10 original VQAv2 answers are $y$, so that $y$ is the single correct answer we hold fixed while editing; 
(3) the answer is not a color, room, side, time or count, which our caption edits change, 
(4) one of the image's five COCO captions contains $y$ and the question's other nouns; this caption is the source caption $c_0$. 
The number of test images in the source and generated datasets are given in \cref{tab:sizes}. 

\paragraph{Models for Benchmarking.} 
We perform experiments on two tasks: image classification and visual question answering (VQA). 
For image classification, to cover a wide range of models spanning different sizes (25.6M to 304M parameters), training protocols (supervised, self-supervised, contrastive, masked image modeling) and different architectures (convolutional networks, and vision transformers, and CLIP-family encoders), we evaluate ResNet-50~\citep{He_2016_CVPR}, ConvNeXt-B~\citep{liu2022convnet}, ViT-B/16~\citep{dosovitskiy2021an}, CLIP-FT B/32~\citep{pmlr-v139-radford21a}, Dinov2-B~\citep{oquab2024dinov}, EVA-02-B~\citep{fang2024eva} as target models, 
along with zero-shot targets CLIP ViT-B/32, OpenCLIP ViT-L/14~\citep{pmlr-v139-radford21a, Cherti_2023_CVPR}, and SigLIP-B/16~\citep{zhai2023sigmoid}. 
For VQA, we evaluate eight models from the Qwen~\citep{qwen36_35b_a3b}, Gemma~\citep{team2026gemma}, Mistral \cite{mistral_small_3_2_25_06}, InternVL~\citep{wang2025internvl3_5}, Muse~\citep{museglimmerteam2026museglimmer30b}, Phi~\citep{abdin2024phi}, and MiniCPM~\citep{hu2024minicpm} families. For models that do not have existing trained checkpoints on the datasets being evaluated, we train them with standard recipes. 

\paragraph{Transfer.} 
We use one surrogate model $F_s$ to create stress-test $\hat{\mathcal{D}}$ and define \textit{transfer} as testing on a target model $F_t$ 
that was not involved in the collection of $\mathcal{\hat{D}}$, thus ensuring that results apply across models and are not tied to the surrogate.
We report the evaluation results on the $\mathcal{\hat{D}}$ obtained from the weakest surrogate but evaluate stronger models in \cref{fig:main_results}, with results with other surrogates reported in the supplement.

\subsection{Results on Visual Recognition}
We evaluate all the models at depths $d=\{1,2,3\}$ and compare with accuracies on the real source images and prior work. 
Moreover, models of different training protocols, different architectures, and different sizes all have significant performance degradation on a dataset optimized only for ResNet-50, with similar findings using other surrogates.
This indicates shared semantic points of failures across surrogates which we analyze in \cref{sec:analysis}.

\paragraph{Comparison with Automated Stress-Tests.}
We compare with a relevant prior approach, LANCE \cite{prabhu2023lance} that uses caption edits and diffusion-based image generation to build stress tests.
For LANCE, we run the released code on source images from every dataset.
Note that even though LANCE does not do hard sample selection using surrogate loss, we still compare with LANCE+our surrogate, keeping the LANCE edit with the highest surrogate loss.
We see in \cref{fig:main_results} that Goldilocks outperforms prior work at depth 1 in most benchmarks, and far outperforms them by depth 3. 
Goldilocks is significantly faster than LANCE, taking 1.68 minutes at d=3 (three rounds of edits) and faster at d=1 and d=2, compared to 19.35 minutes for LANCE's single-round edit.

\begin{figure*}
    \centering
    \includegraphics[width=\linewidth]{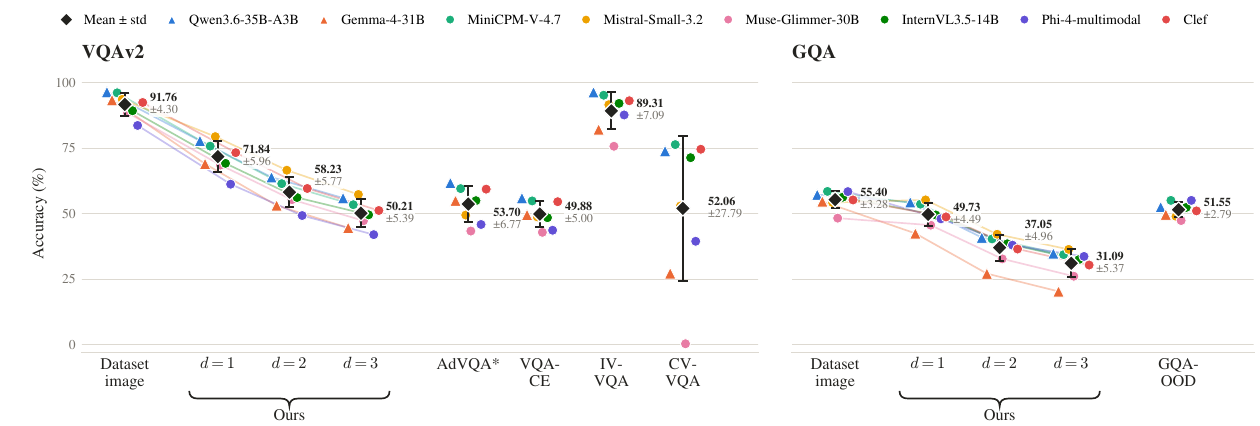}
    \caption{Comparing the performance of vision-language models on existing benchmark datasets, prior work on VQA robustness evaluation, and our Goldilocks Zone test sets generated at depths $d=1,2,3$. We use Gemma-4 12B as a surrogate and test the performance of stronger SOTA models. 
    Goldilocks datasets are challenging for all target models on each benchmark with accuracy falling as depth increases.}
    \label{fig:vqa_results}
\end{figure*}

\paragraph{Comparisons with human-compiled evaluation sets.}
In this experiment, we compare Goldilocks' performance with other human-compiled datasets. A human-compiled dataset is a dataset that has some level of human involvement in its curation, from being collected entirely by humans (such as ImageNet-A) to humans being a validity check to an automated generation process (such as ImageNet-D). We select these datasets for their representation of hard natural images, OOD settings, varying backgrounds, and spurious correlations. We report results on ImageNet-A, -R, -D, -Sketch, and -V2, ObjectNet, and CounterAnimal ~\cite{counteranimal, hendrycks2021many, hendrycks2021natural, recht2019imagenet, wang2019learning, barbu2019objectnet, zhang2024imagenet}. We score each on its own classes and compare performance with our created dataset on the same classes, shown in \cref{fig:human_comp}. 
Out of the selected datasets, the targets' performance drops to 27.13\% on ImageNet-A and 19.69\% on ImageNet-D, because of their being created with data points curated/selected to degrade model performance. However, Goldilocks images on those classes cause further degradation to 22.62\% and 11.77\% respectively at depth 3, showcasing Goldilocks' difficulty. On the other datasets, Goldilocks often outperforms at depth 1 or 2 because they are not collected to degrade model performance.  

\begin{figure}
    \centering
    \includegraphics[width=\linewidth]{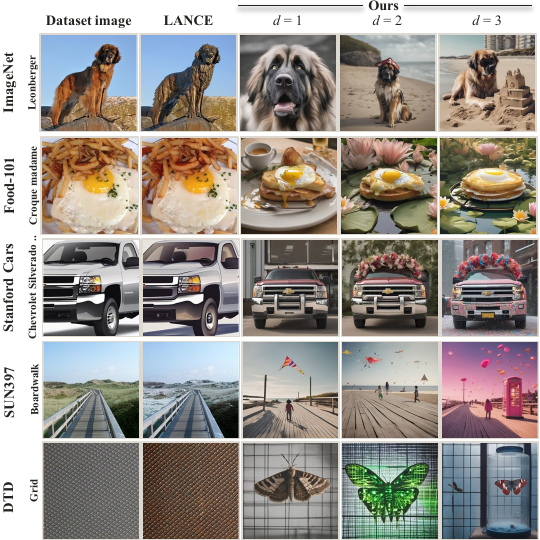}
    \caption{Qualitative examples of images generated by LANCE and Goldilocks, on five image classification benchmarks.}
    \label{fig:qual_img}
\end{figure}

\paragraph{Qualitative Examples.}
\cref{fig:qual_img} compares the original image, LANCE generation, and Goldilocks generations at each depth. The edits produced by Goldilocks meaningfully move away from the source image, adding new objects and introducing confounding factors along the different axes, while LANCE edits do not change significantly. For example, in the Leonberger edits, we see that Goldilocks produces images of the dog in various positions, while adding context in the background (the beach) and objects that interact with the ground truth object (hat, sunglasses). Similar changes can be observed across the other examples as well, with Goldilocks preserving the ground truth labels while making edits to the image. In the case of DTD, the texture is applied to other objects to ensure that the ground truth answer is still visible in the edit, while making the image itself challenging.

\subsection{Results on Visual Question Answering}

We evaluate vision-language models on open-ended VQA on 2 datasets $\times$ 2 surrogates $\times$ 8 targets, once again covering different training protocols and different scales of architectures. 
We report results using the weakest surrogate (Gemma-4-12B) in our generation pipeline, but evaluate stronger models.
We also compare to other VQA robustness benchmarks, AdVQA, VQA-CE, IV-VQA, GQA-OOD and CV-VQA~\citep{9710147,9156407,sheng2021human,kervadec2021roses}. \cref{fig:vqa_results} shows a decrease in accuracy, falling further at higher depths. 

\begin{figure}
    \centering
    \includegraphics[width=\linewidth]{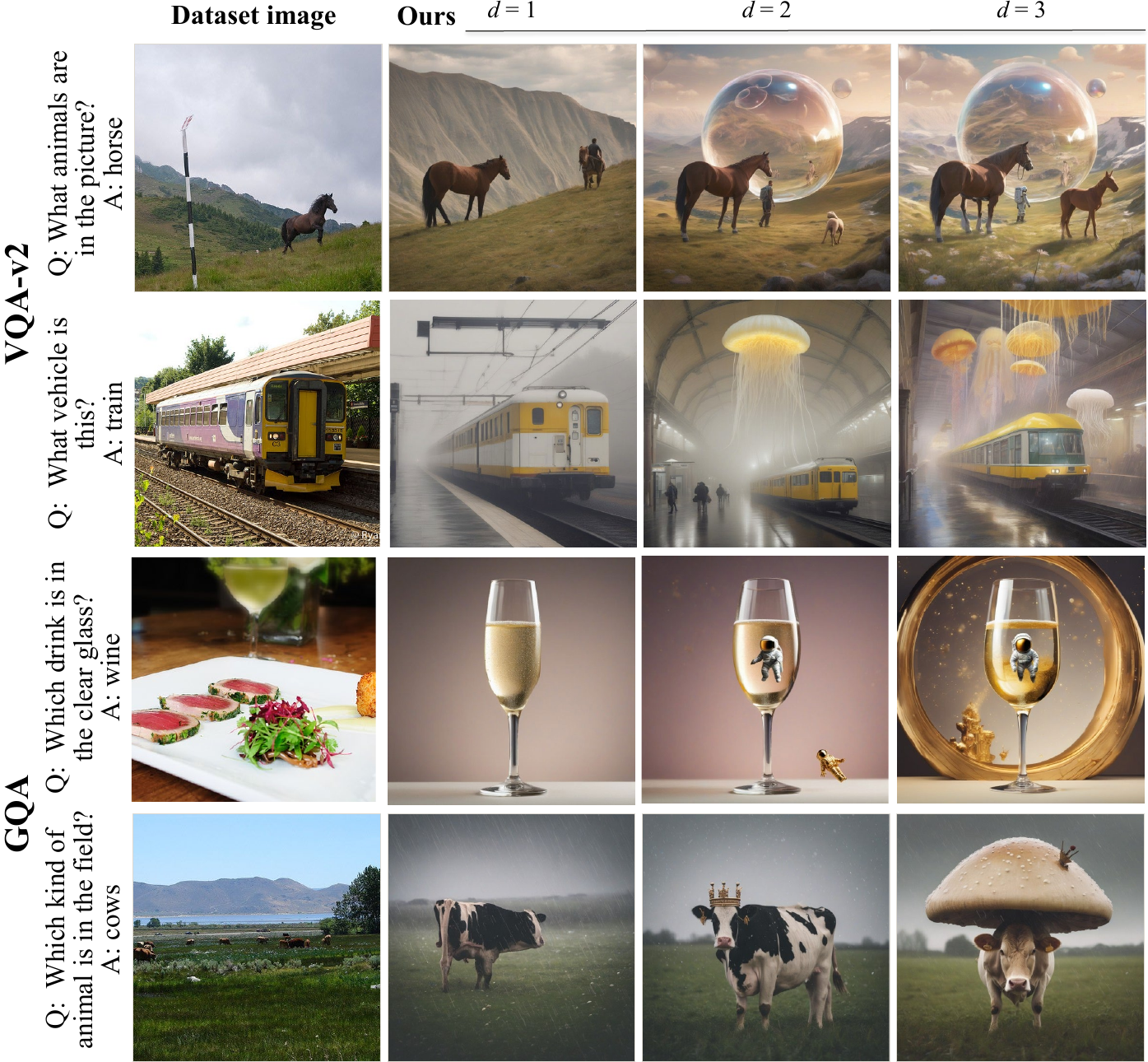}
    \caption{Qualitative examples comparing images from VQA-v2 and GQA with our Goldilocks images.}
    \label{fig:qual_vqa}
\end{figure}

\paragraph{Qualitative examples} 
in \cref{fig:qual_vqa} show meaningful changes (a horse in the mountains) made to the source image, while ensuring the ground truth answer is preserved in the image. Our method also introduces added objects like a bubble behind the horse or a crown on the cow which does not change the answer to the question ``which animal''.
\vspace{-4pt}
\section{Analysis}
\vspace{-4pt}
\label{sec:analysis}
\begin{figure}[t]
    \centering
    \includegraphics[width=\linewidth]{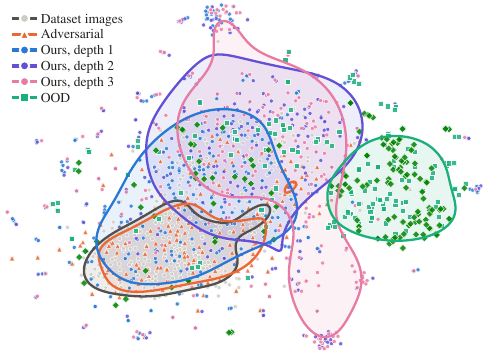}
    \caption{t-SNE plot comparing distributions of original, OOD, adversarial, and Goldilocks images. Adversarial images are close to the source distribution, and OOD images are far away.
    Our generated images gradually move away from the source with increasing depth $d$, but remain between adversarial and OOD. 
    Each image is represented by its Inception-v3 features minus mean feature of real photos of its class. Outlines enclose the densest half of each group. 
    }
    \label{fig:tsne}
\end{figure}

\paragraph{Goldilocks images are in the middle.}
To understand where the image distribution generated by the Goldilocks approach lies in comparison to adversarial examples and out-of-distribution (OOD) data, we plot a t-SNE map of real ImageNet images, per-image adversarial images ($\ell_\infty{=}4/255$ PGD attack on ResNet-50), images generated by our method at depths 1 through 3 (with ResNet-50 surrogate), and OOD datasets (ImageNet-R and ImageNet-Sketch) based on ImageNet. 
The adversarial distribution is nearly identical to real images, the OOD datasets are far apart from the real images.
Images generated in the Goldilocks Zone are close to real images at depth 1, and gradually move away at depths 2 and 3, lying between real and OOD distributions. 

\begin{table}[t]
\caption{Analysis across four target ImageNet classifiers (ResNet-50, ViT-B, ConvNeXt, and zero-shot CLIP) when using (a) different text-to-image generators, and (b) different perplexity scoring models.
We use the strongest image generation and perplexity models.
}
\label{tab:swaps_combined}
\centering
\small
\begin{subtable}[t]{0.48\linewidth}
    \centering
    \caption{Impact of text-to-image generator on the mean top 1 accuracy (\%).}
    \label{tab:t2i_model}
    \resizebox{\linewidth}{!}{
    \begin{tabular}{@{}lccc@{}}
    \toprule
    Generator $G$          & $d=1$ & $d=2$ & $d=3$ \\
    \midrule
    SDXL             & 77.0  & 49.5  & 24.0  \\
    SD 1.5                 & 69.3  & 45.8  & 21.8  \\
    PixArt-$\Sigma$        & 61.3  & 40.5  & 21.3  \\
    \bottomrule
    \end{tabular}
    }
\end{subtable}
\hfill
\begin{subtable}[t]{0.48\linewidth}
    \centering
    \caption{Impact of the perplexity scorer on the mean top 1 accuracy (\%).}
    \label{tab:ppl_model}
    \resizebox{\linewidth}{!}{
    \begin{tabular}{@{}lccc@{}}
    \toprule
    PPL Scorer      & $d=1$ & $d=2$ & $d=3$ \\
    \midrule
    LLaMA-3.1-8B     & 74.2  & 48.5  & 31.6  \\
    GPT-2                  & 73.6  & 47.7  & 31.5  \\
    Qwen2.5-1.5B           & 74.0  & 48.2  & 31.6  \\
    \bottomrule
    \end{tabular}
    }
\end{subtable}
\end{table}

\paragraph{Effects of Choice of Image Generator and Perplexity Scorer.} To demonstrate the robustness of our method, we swap out our image generation model (SDXL) for alternatives (SD 1.5~\citep{Rombach_2022_CVPR}, and PixArt-$\Sigma$~\citep{DBLP:conf/eccv/ChenGXWYRWLLL24}), and measure its impact on downstream performance in \cref{tab:t2i_model}.
Similarly, we compare the impact of perplexity scoring language models (LlaMA-3.1-8B, GPT-2~\citep{radford2019language}, and Qwen2.5-1.5B~\citep{qwen2025qwen2}) in \cref{tab:ppl_model}.
Captions selected by different perplexity scorers all cause similar performance drops across the test suite; images generated by different T2I models are more challenging as depth increases.
These results demonstrate our results are agnostic to modules used in the pipeline.

\paragraph{Human-Detector Agreement on Generated Images.} 
We sample 700 images uniformly from across three depths and ask human participants to report via multi-choice options whether the image contains (A) the ground truth label (B) the model's prediction, (C) both, or (D) neither. We collect responses from 7 respondents, with 100 images per respondent. 
Comparing the results of the human respondents to detector results, we note that the recall is $84\%$, thus establishing that the detector is a suitable proxy for human validity checks.
Details of this study are in the supplementary material.

\begin{figure}[t]
    \centering
    \includegraphics[width=\linewidth,trim={0 6 0 6}, clip]{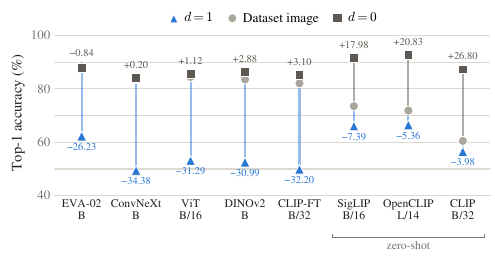}
    \caption{Model performance does not degrade beyond $1\%$ on unperturbed synthetic renditions of ImageNet classes ($d=0$). 
    However, Goldilocks images (with $d=1$ linguistic perturbations) are challenging for models with large accuracy drop. 
    }
    \label{fig:synth}
\end{figure}

\paragraph{Drop in accuracy is not due to synthetic nature of images.}
We investigate whether the drop in accuracy can be attributed to the synthetic nature of images or to the actual difficulty posed by Goldilocks images.
We generate synthetic images by using the prompt template ``an photo of a {\tt y}'' as input to a text-to-image diffusion model \cite{podellsdxl}, where {\tt y} are ImageNet class labels. 
From \cref{fig:synth} we observe that accuracy on these synthetic renditions (without our perturbations) either remains consistent, or in some cases improves, whereas accuracy on our images (with Goldilocks Zone perturbations) decreases. 
This establishes clearly that our challenging semantic edits and not simply the synthetic nature of images is why models fail in the Goldilocks Zone.

\begin{figure}[t]
    \centering
    \includegraphics[width=\linewidth,trim={0 6 0 6}, clip]{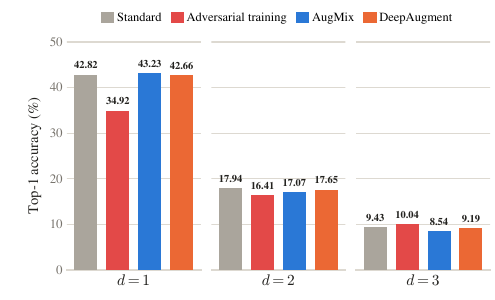}
    \caption{Comparing the performance of four training protocols for ResNet-50: standard, adversarially trained~\cite{DBLP:conf/nips/CroceASDFCM021}, and domain shift robust approaches~\citep{hendrycks2021many, hendrycks*2020augmix} on Goldilocks images. 
    We see that prior robust training approaches are ineffective against Goldilocks images, with no significant improvement over standard training.
    }
    \label{fig:robust}
\end{figure}

\paragraph{Adversarial Training and Shift Robust Models do not mitigate the challenge of Goldilocks Images.} \label{semantic_failures}
Prior work has already identified that models trained for adversarial robustness do not mitigate domain shift and vice versa~\citep{gokhale-etal-2022-generalized, moayeri2022explicit}. 
We evaluate adversarially trained models~\citep{DBLP:conf/nips/CroceASDFCM021} and shift-robust models~\citep{hendrycks2021many, hendrycks*2020augmix} on Goldilocks images, 
and find that neither is effective, with both training protocols having a low accuracy across all depths of Goldilocks images. 
Our work therefore reveals the need to develop algorithms that must be robust across the spectrum of stress-testing datasets, and not just catered to one dimension of robustness. 

\begin{table}[t]
\caption{Top 5 common edits preferred and avoided by all six surrogates while selecting the hardest variant. \emph{Classes picking edit} is the number of ImageNet classes with at least one source image with the edit proposed, \emph{Consensus hard images} is the number of images with that edit that all six surrogates selected as the hardest variant, and \emph{Accuracy $\Delta$} is how many points the accuracy of the three zero-shot models drops or rises on those images, compared with the other variants of the same source image at the same depth.}
\centering
\footnotesize
\begin{tabular}{@{}clrrr@{}}
\toprule
& Edit & \shortstack[l]{\# Classes \\picking edit} & \shortstack[l]{\# Consensus \\ hard images} & \shortstack[l]{Accuracy $\Delta$ \\ (\%)} \\
\midrule
\multirow{5}{*}{\rotatebox[origin=c]{90}{\shortstack[c]{Commonly \\ Selected}}} 
& hot-air balloon   & 848 & 569 & $-$11.9 \\
& room              & 821 & 353 &  $-$2.2 \\
& telephone booth   & 648 & 166 &  $-$9.4 \\
& chandelier        & 799 & 114 &  $-$2.4 \\
& ceiling           & 575 &  78 &  $-$3.0 \\
\midrule
\multirow{5}{*}{\rotatebox[origin=c]{90}{\shortstack[c]{Commonly \\ Avoided}}} 
& sidewalk          &  39 &   1 &  $+$0.3 \\
& headphone         & 142 &   1 &  $+$4.4 \\
& top hat           & 171 &   1 &  $-$0.2 \\
& suitcase          & 190 &   2 &  $+$2.5 \\
& salt flat         & 152 &   2 &  $+$8.3 \\
\bottomrule
\end{tabular}
\label{tab:shared_objects}
\end{table}

\paragraph{Common Semantic Failures Across Surrogate/Target Models.} 
To study common semantic failures across our target models, we run two experiments on ImageNet with six surrogates: (1) whether common edits are selected as the hardest image by all surrogates regardless of source image; and (2) how three zero-shot models that are never used as surrogates perform on these selected images. In \cref{tab:shared_objects}, we see that these commonly selected edits correspond to declining accuracy of up to 11.9\% on the target models. Similarly, commonly avoided edits cause accuracy of the target models to rise. This indicates that shared vulnerabilities exist across these surrogates, despite their being trained with different protocols and having different architectures.
\vspace{-4pt}
\section{Conclusion}
\vspace{-4pt}
We present the Goldilocks Zone, a ``just right'' testing ground for downstream models, and an automated, data-agnostic framework that generates progressively challenging, linguistically plausible visual edits to explore the testing ground. We evaluate several state-of-the-art models on stress tests generated by our method and show that their performance degrades sharply on Goldilocks edits. Further, we show that models share semantic vulnerabilities that can be exploited to degrade performance further. Together, these results show that stress-testing in the Goldilocks Zone reveals an underexplored failure mode of state-of-the-art models and opens a pathway to systematically exploring plausible visual edits.
\clearpage
\appendix

\maketitlesupplementary
\section{Pseudo-code}
The pseudo-code of the main Goldilocks method is given here. 

\begin{algorithm}[h]
\small
\renewcommand{\algorithmicrequire}{\textbf{Input:}}
\renewcommand{\algorithmicensure}{\textbf{Output:}}
\caption{Goldilocks Zone Stress Test Generation}
\label{alg:goldi}
\begin{algorithmic}[1]
    \Require image $x$ with ground-truth label $y$; initial caption $c_0$ containing $y$; perturber $P$; perplexity threshold $\tau$; generator $G$; detector $D$; surrogate $F_s$; maximum depth $T$
    \Ensure maximally challenging image $\hat{x}_d$ and its caption $c_d$ for every depth $d \leq T$ reached
    \State $A \gets \emptyset$ \Comment{locked edits}
    \For{$d = 1, \dots, T$}
        \State $\hat{\mathcal{C}} \gets P(c_{d-1}, A)$ \Comment{Subject, Adjective, Context; keep $y$ and $A$}
        \State $\mathcal{C}_\mathrm{Goldilocks} \gets \{\hat{c} \in \hat{\mathcal{C}} \mid \mathrm{PPL}(\hat{c}) < \tau\}$ \Comment{main paper, Eq.~(1)}
        \State $\mathcal{X} \gets \{G(\hat{c}) \mid \hat{c} \in \mathcal{C}_\mathrm{Goldilocks}\}$ \Comment{one image per caption}
        \State $\mathcal{X}_\mathrm{verified} \gets \{x \in \mathcal{X} \mid y \in D(x)\}$ \Comment{main paper, Eq.~(2)}
        \If{$\mathcal{X}_\mathrm{verified} = \emptyset$} \State \textbf{break} \EndIf
        \State $\hat{x}^\ast \gets \argmax_{x \in \mathcal{X}_\mathrm{verified}} \mathcal{L}\big(F_s(x), y\big)$, where $\hat{x}^\ast = G(\hat{c}^\ast)$ \Comment{main paper, Eq.~(3)}
        \State $c_d \gets \hat{c}^\ast$; \quad $\hat{x}_d \gets \hat{x}^\ast$ \Comment{bootstrap}
        \State $A \gets A \cup \{\text{the element added in } c_d\}$ \Comment{locked edit at depth $d$}
    \EndFor
\end{algorithmic}
\end{algorithm}

\begin{figure}[h]
    \centering
    \includegraphics[width=\linewidth]{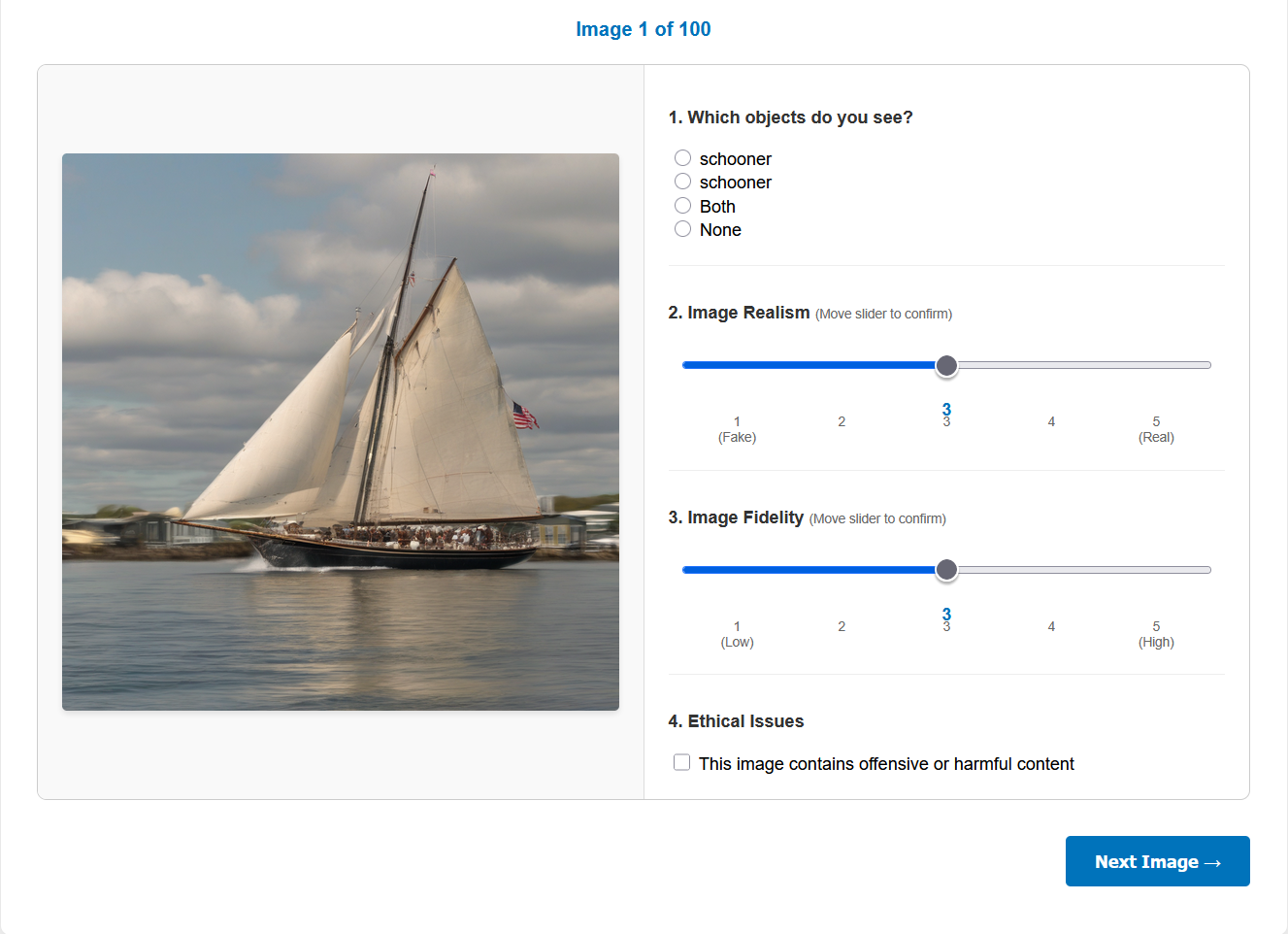}
    \caption{This image shows the interface of the human annotator study we conducted. The humans would see the generated image, and have a multiple choice question containing the ground truth object, the predicted object, both, and neither as options. The ``both'' option meant that the human could see both the predicted object and the ground truth object in the image. 
    }
    \label{fig:human_interface}
\end{figure}

\begin{figure}[t]
    \centering
    \includegraphics[width=\linewidth]{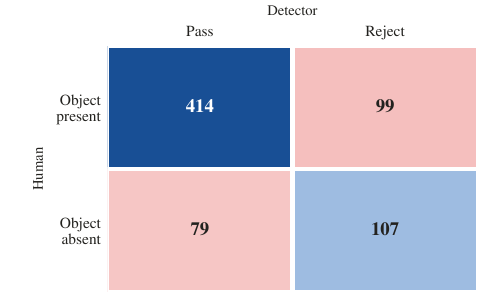}
    \caption{Human-detector agreement on the set of 699 images human respondents answered questions about. We see that the detector achieves an 84\% precision, showing that the detector serves as a suitable proxy for human testing. 
    }
    \label{fig:confusion_matrix}
\end{figure}

\section{Alternate Surrogate Results -- Image Classification}

\begin{figure*}[t]
    \centering
    \includegraphics[width=\linewidth,trim={0 6 0 0}, clip]{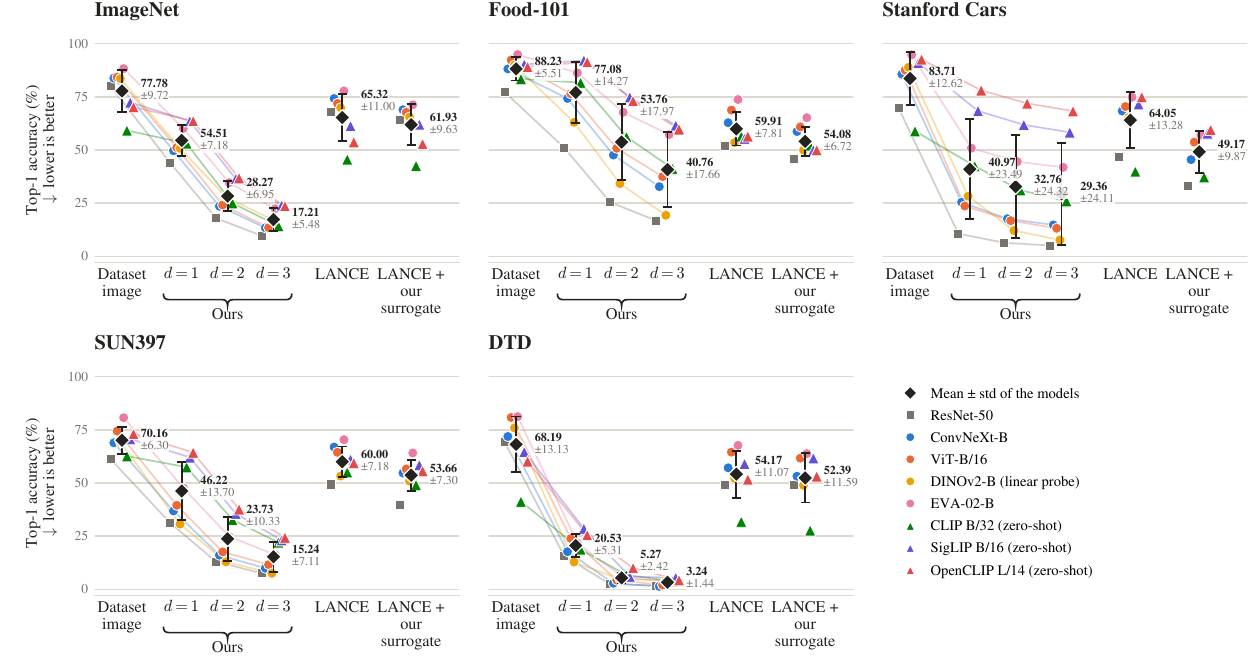}
    \caption{
    Comparing the performance of image classification models on existing benchmark datasets with LANCE \cite{prabhu2023lance} and our Goldilocks Zone test sets generated at depths $d=1, 2, 3$.
    We use CLIP-FT B/32 as an alternative surrogate and test the performance of the other target models (circles: supervised; triangles: zero-shot). 
    Goldilocks datasets are challenging for all target models on each benchmark, with accuracy declining at higher depth of the perturbation.
    }
    \label{fig:surr_1}
\end{figure*}

\begin{figure*}[t]
    \centering
    \includegraphics[width=\linewidth,trim={0 6 0 0}, clip]{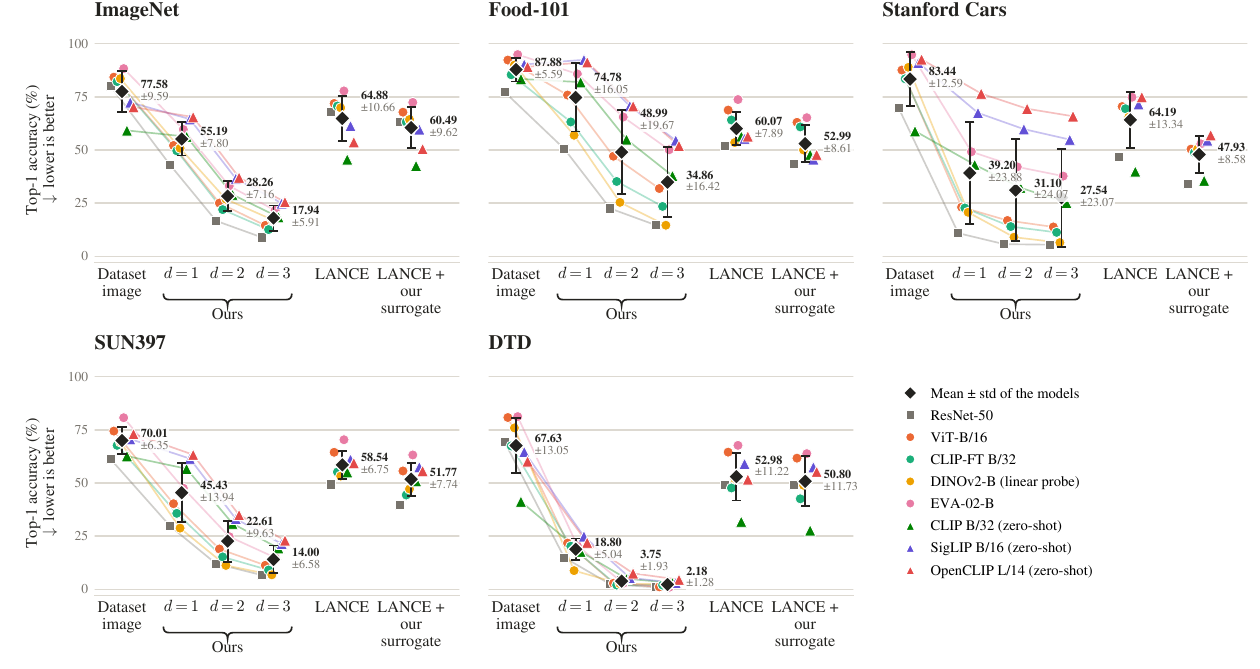}
    \caption{
    Comparing the performance of image classification models on existing benchmark datasets with LANCE \cite{prabhu2023lance} and our Goldilocks Zone test sets generated at depths $d=1, 2, 3$.
    We use ConvNeXt-B as an alternative surrogate and test the performance of the other target models (circles: supervised; triangles: zero shot). 
    Goldilocks datasets are challenging for all target models on each benchmark, with accuracy declining at higher depth of the perturbation.
    }
    \label{fig:surr_2}
\end{figure*}

\begin{figure*}[t]
    \centering
    \includegraphics[width=\linewidth,trim={0 6 0 0}, clip]{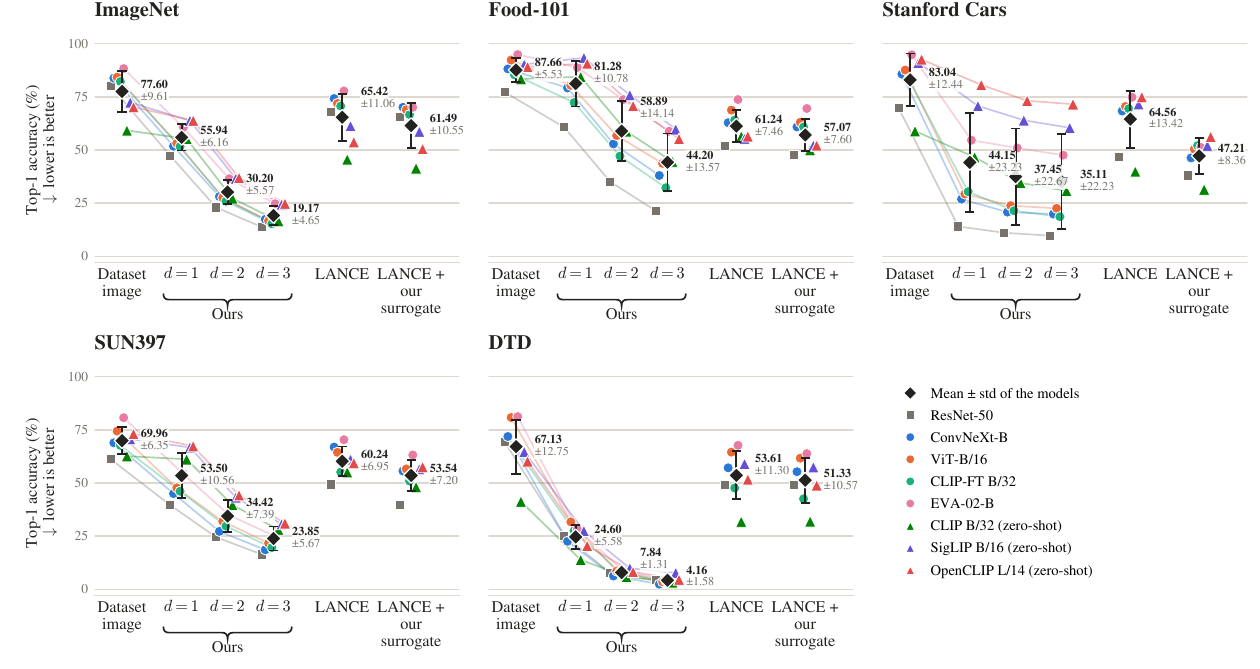}
    \caption{
    Comparing the performance of image classification models on existing benchmark datasets with LANCE \cite{prabhu2023lance} and our Goldilocks Zone test sets generated at depths $d=1, 2, 3$.
    We use Dino-v2-B as an alternative surrogate and test the performance of the other target models (circles: supervised; triangles: zero shot). 
    Goldilocks datasets are challenging for all target models on each benchmark, with accuracy declining at higher depth of the perturbation.
    }
    \label{fig:surr_3}
\end{figure*}

\begin{figure*}[t]
    \centering
    \includegraphics[width=\linewidth,trim={0 6 0 0}, clip]{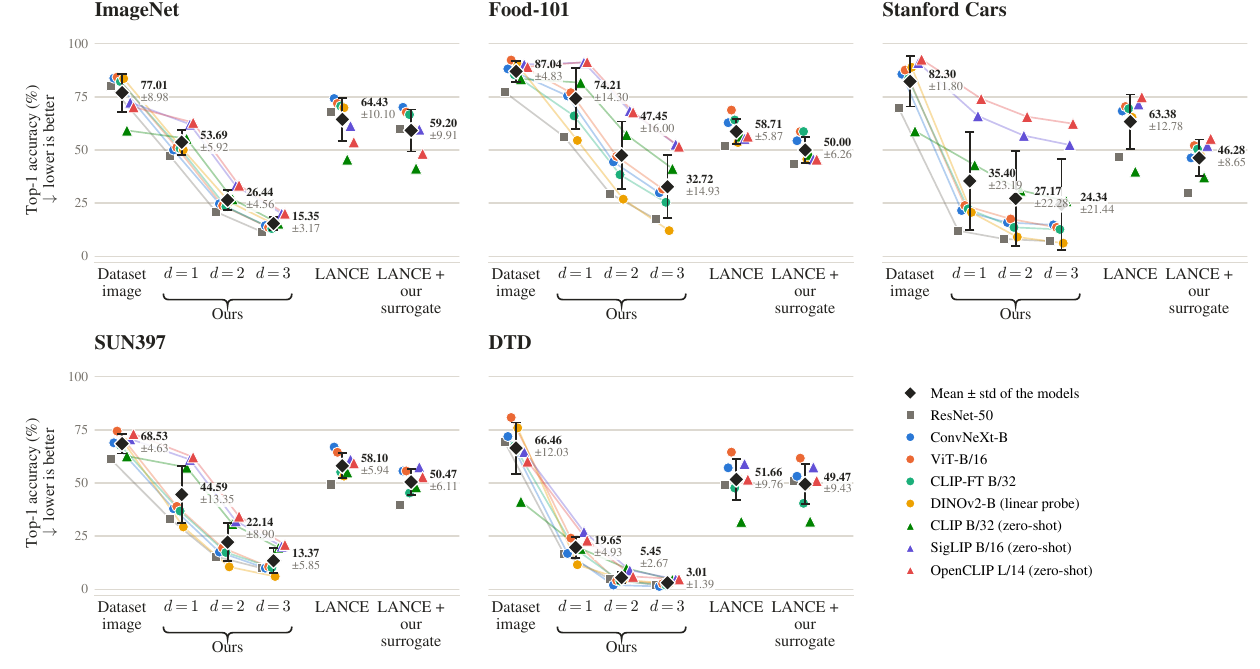}
    \caption{
    Comparing the performance of image classification models on existing benchmark datasets with LANCE \cite{prabhu2023lance} and our Goldilocks Zone test sets generated at depths $d=1, 2, 3$.
    We use Eva-02-B as an alternative surrogate and test the performance of the other target models (circles: supervised; triangles: zero shot). 
    Goldilocks datasets are challenging for all target models on each benchmark, with accuracy declining at higher depth of the perturbation.
    }
    \label{fig:surr_4}
\end{figure*}

\begin{figure*}[t]
    \centering
    \includegraphics[width=\linewidth,trim={0 6 0 0}, clip]{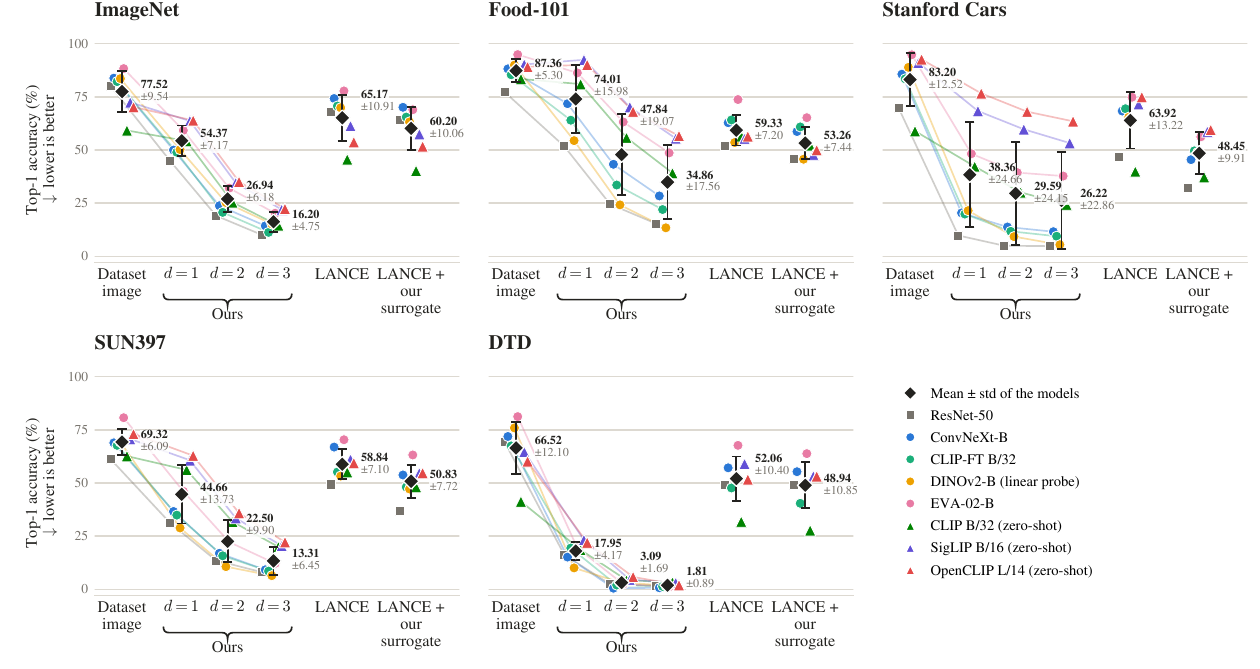}
    \caption{
    Comparing the performance of image classification models on existing benchmark datasets with LANCE \cite{prabhu2023lance} and our Goldilocks Zone test sets generated at depths $d=1, 2, 3$.
    We use ViT-B/16 as an alternative surrogate and test the performance of the other target models (circles: supervised; triangles: zero shot). 
    Goldilocks datasets are challenging for all target models on each benchmark, with accuracy declining at higher depth of the perturbation.
    }
    \label{fig:surr_5}
\end{figure*}

We report the results of the five other surrogates we use for the image classification hard image search in \cref{fig:surr_1}, \cref{fig:surr_2}, \cref{fig:surr_3}, \cref{fig:surr_4}, and     \cref{fig:surr_5}. There is a drop in accuracy across all surrogates, showing that the effect observed in the main results due to Goldilocks images is not due to a specific surrogate. All six surrogates, which span convolutional, transformer, self-supervised and CLIP-family classifiers, produce the same drop, so the effect does not depend on the choice of surrogate.

 We report two variants of LANCE-- one accepted edit chosen at random, and one chosen specifically by a surrogate. 

\section{Alternate Surrogate Results -- Visual Question Answering}

\begin{figure*}[h]
    \centering
    \includegraphics[width=\linewidth]{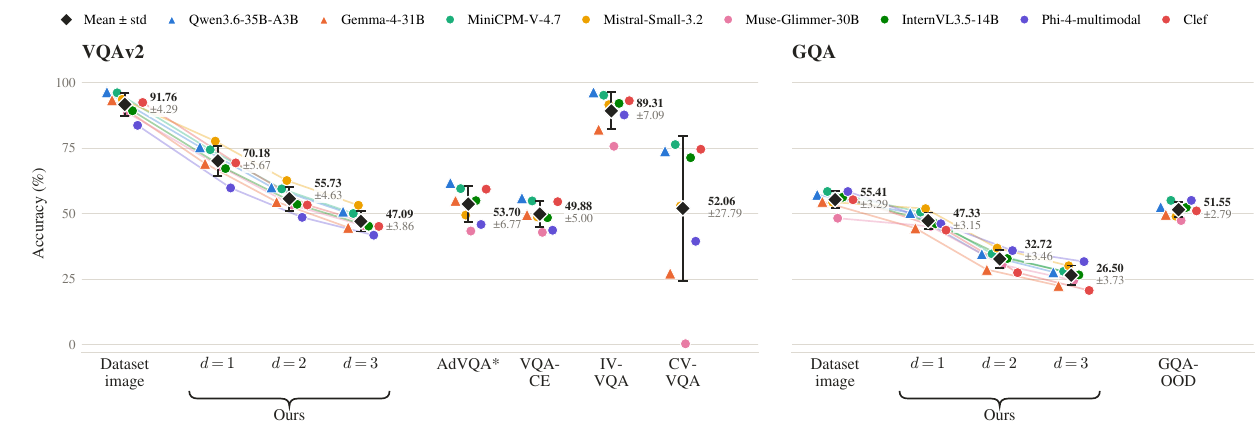}
    \caption{
    Comparing the performance of vision-language models on existing benchmark datasets, prior work on VQA robustness evaluation, and our Goldilocks Zone test sets generated at depths $d=1,2,3$. We use Qwen3.8-27B as a surrogate and test the performance of the other target models. 
    Goldilocks datasets are challenging for all target models on each benchmark with accuracy falling as depth increases.
    }
    \label{fig:surr_vqa}
\end{figure*}

Similarly, we report the results of the other surrogate we use for the visual question answering task in \cref{fig:surr_vqa}. As observed in the image classification task, there is a drop in accuracy observed in every surrogate setting, indicating that Goldilocks images causes issues in downstream models.  

\section{Human Study}

We report the human-detector agreement in this section. The human study was conducted with 7 participants, each having a set of 100 images uniformly sampled from each depth $d=1,2,3$. These images were generated without the detector as a part of the pipeline, to fully understand how the detector affects the pipeline. The total number of images were 699. The human annotators were asked what they saw in each image, with a multiple choice question selection of the a) the ground truth; b) the model prediction, c) both, and d) neither being offered. The interface is given in \cref{fig:human_interface}. We pass the same set of images through the detector to measure human-detector agreement. We present the human-detector agreement results in \cref{fig:confusion_matrix}. We see that out of the 699 images, humans and the detector agree on 521 of them. 99 of the images are rejected by the detector while having the ground truth object in them, and 79 are passed by the detector while not having the ground truth object in them.

\begin{figure*}[h]
    \centering
    \includegraphics[width=\linewidth,trim={0 6 0 0}, clip]{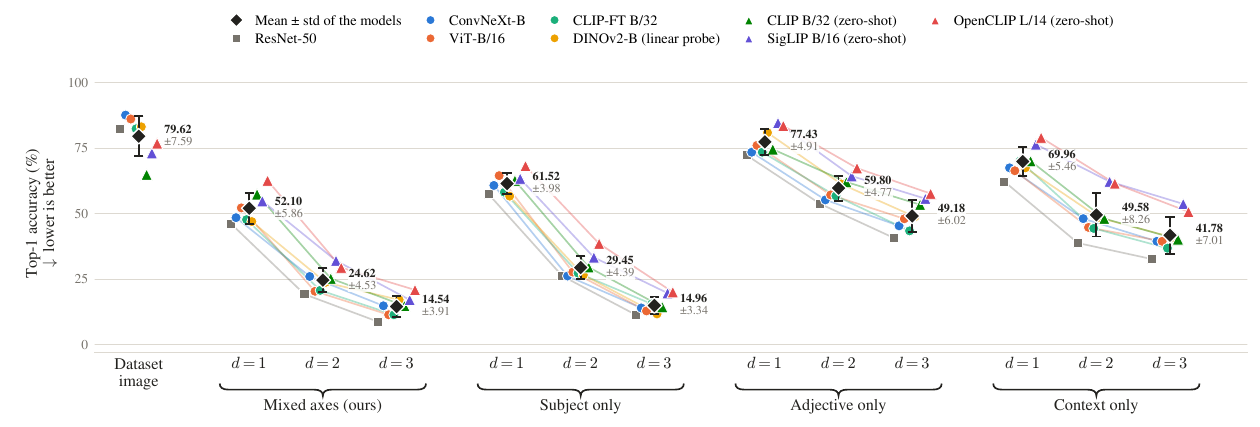}
    \caption{
    We study the effects of each axis on accuracy by isolating searches that use only one axis in the larger set. We use EVA-02-B as the surrogate model, and test the accuracy drops across 268 real images. We see that the subject axis has the most effect on accuracy.   
    }
    \label{fig:axis_breakdown}
\end{figure*}

\section{Axes Breakdown}

To better understand the different axes and their impacts on the iterative search for hard images, we break down the hardest images for one surrogate by axis and study the effects of each axis on accuracy. We present these results in \cref{fig:axis_breakdown}. We see that the subject axis causes the most accuracy drops, almost mimicking the accuracy drop of the mixed axes. Contrary to the subject axes, the adjective and context axes cause smaller drops.

\section{Perturber Prompt}

We provide our perturber prompts for depths $d = 1,2,3$ here. We use a system message with a modifiable \{task\} parameter that is changed depending on depth.

\begin{promptbox}{Perturber prompt (system message; one call returns 10 variants)}
You are an expert adversarial stress-tester for computer vision models.

CRITICAL RULE 1: Your new caption MUST contain the EXACT string '|{gt}|'. NEVER replace it with a synonym, a hypernym, or a different object.
CRITICAL RULE 2: You MUST preserve all previous edits exactly. Here is the list of previously added elements you cannot remove or alter: [|{locked}|]

Ground Truth Object: |{gt}|
Base Caption: |{base}|

Task: |{task}|

Format your response EXACTLY as a JSON object with a key 'variants' mapping to a list of 10 objects.
Example:
{"variants": [{"caption": "A photo of a |{gt}| in a neon spacesuit", "newly_added_element": "in a neon spacesuit"}]}
\end{promptbox}

\begin{promptbox}{\{task\} at depth $d=1$ (\{locked\} = None)}
Generate 10 variations. Each variation must modify exactly ONE axis, and across the 10 you must use each axis at least three times:
  - Subject: introduce another entity that interacts with or occupies the scene alongside the object
  - Adjective: change a visual property of the object itself
  - Context: change the setting, surroundings, weather, or time of day
Prefer edits that are unusual yet physically plausible. Do not produce ten edits of the same axis.
\end{promptbox}

\begin{promptbox}{\{task\} at depth $d \geq 2$}
Add one new, highly distinct element (Subject, Adjective, or Context) to the Base Caption to further distract the scene.
\end{promptbox}

\begin{promptbox}{VQA question prompt (surrogate and target VLMs)}
|{q}|
Answer the question using a single word or phrase.
\end{promptbox}

\clearpage
{
    \small
    \bibliographystyle{ieeenat_fullname}
    \bibliography{main}
}

\end{document}